\documentclass{article}
\usepackage{spconf,amsmath,graphicx,hyperref}

\usepackage[utf8]{inputenc}
\usepackage{multirow}
\usepackage{bbding}
\usepackage{amsthm}

\newtheorem{definition}{Definition}
\usepackage{amssymb}
\usepackage{booktabs}

\title{Robust Uncertainty Estimation under Distribution Shift via Difference Reconstruction}
\name{Xinran Xu$^{1}$ \qquad Li Rong Wang$^{2}$ \qquad Xiuyi Fan$^{1,2,3}$*\thanks{*Corresponding author. This study was supported by the Ministry of Education, Singapore (Grant ID: RS15/23).}}
\address{$^{1}$Lee Kong Chian School of Medicine, Nanyang Technological University, Singapore\\
$^{2}$College of Computing and Data Science, Nanyang Technological University, Singapore \\
$^{3}$Centre for Medical Technologies \& Innovations, National Health Group, Singapore \\
xinran007@e.ntu.edu.sg, lirong002@e.ntu.edu.sg, xyfan@ntu.edu.sg}

\begin{document}
\ninept         
\maketitle
\begin{abstract}
Estimating uncertainty in deep learning models is critical for reliable decision-making in high-stakes applications such as medical imaging. Prior research has established that the difference between an input sample and its reconstructed version produced by an auxiliary model can serve as a useful proxy for uncertainty. However, directly comparing reconstructions with the original input is degraded by information loss and sensitivity to superficial details, which limits its effectiveness. In this work, we propose Difference Reconstruction Uncertainty Estimation (DRUE), a method that mitigates this limitation by reconstructing inputs from two intermediate layers and measuring the discrepancy between their outputs as the uncertainty score. To evaluate uncertainty estimation in practice, we follow the widely used out-of-distribution (OOD) detection paradigm, where in-distribution (ID) training data are compared against datasets with increasing domain shift. Using glaucoma detection as the ID task, we demonstrate that DRUE consistently achieves superior AUC and AUPR across multiple OOD datasets, highlighting its robustness and reliability under distribution shift. This work provides a principled and effective framework for enhancing model reliability in uncertain environments.
\end{abstract}
\section{Introduction}
Deep learning models are well known for producing overly confident predictions, often failing to reflect the true level of uncertainty. This limitation undermines their reliability, particularly when the test distribution deviates from the training data \cite{yu2024discretization}. Such distributional shifts are common in real-world deployments, where factors like changes in data acquisition, patient demographics, or imaging devices lead to out-of-distribution (OOD) samples. In safety-critical domains such as healthcare \cite{zou2023review,liang2022advances} and finance \cite{gawlikowski2023survey}, accurately quantifying uncertainty is therefore crucial for identifying misclassifications and detecting cases where models are likely to fail.

\begin{figure}[!htb]
  \centering
  \includegraphics[width=\linewidth]{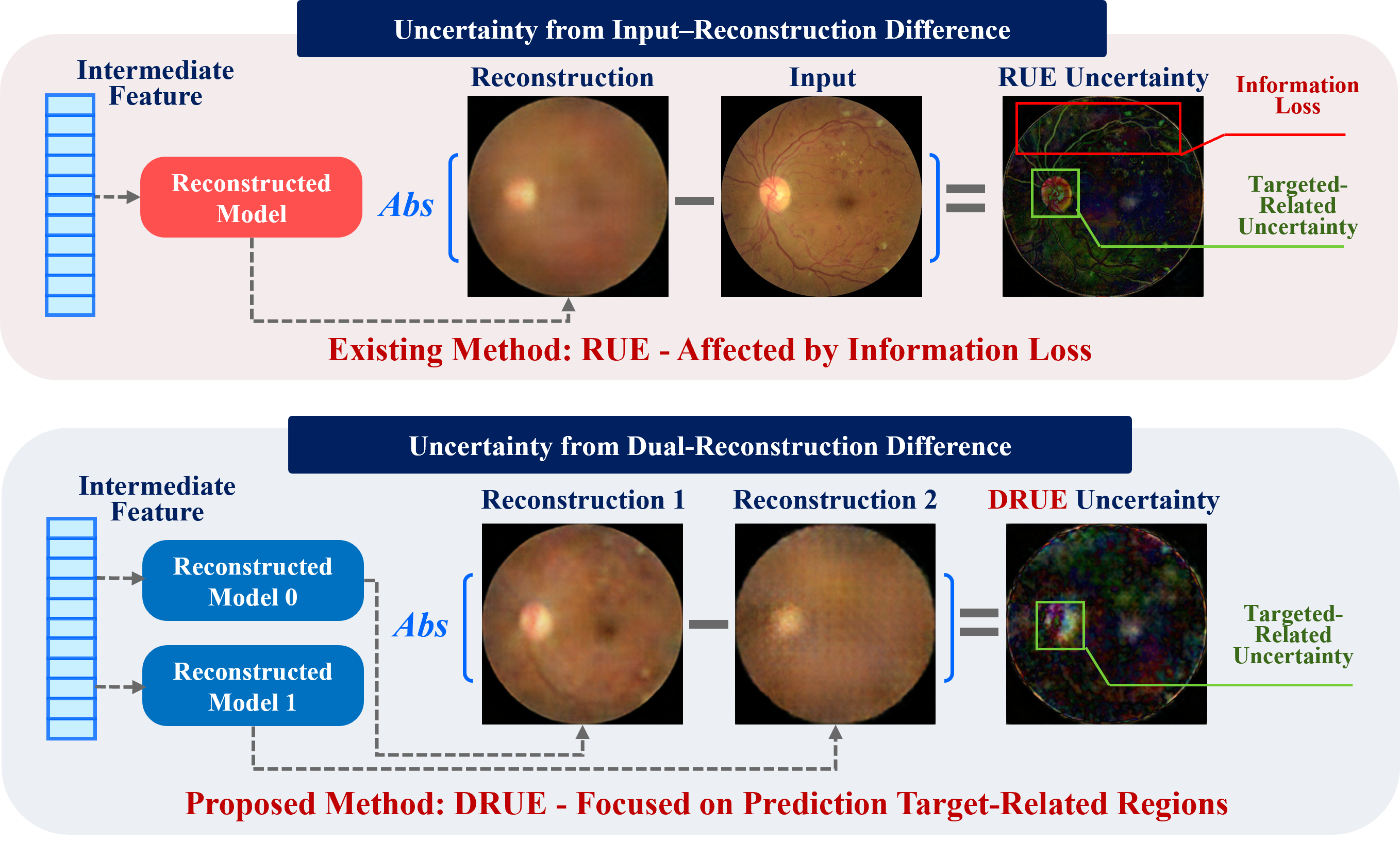}
  \caption{Illustration of reconstruction-based uncertainty estimation in glaucoma detection.
(Top) Conventional input–reconstruction difference often highlights irrelevant regions, such as blood vessels, due to information loss. (Bottom) DRUE compares two reconstructions from different intermediate layers, reducing this effect and focusing uncertainty on clinically meaningful regions, such as the optic disc and cup (the bright, round region).}
  \label{fig:intro2}
\end{figure}

To address this, uncertainty estimation is becoming a major research direction, with existing methods broadly grouped into three categories. (1) Auxiliary estimators compute uncertainty scores without altering the predictive model, such as softmax-based entropy or reconstruction-driven measures \cite{hornauer2022gradient, wang2023uncertainty}. (2) Bayesian and Dirichlet-based models capture epistemic and aleatoric uncertainty through parameter distributions or evidential priors, though they often require substantial retraining or structural changes \cite{blundell2015weight,kendall2017uncertainties, maddox2019simple, sensoy2018evidential, shen2023post}. (3) Ensemble methods combine predictions from multiple models or stochastic passes, producing strong estimates at the expense of heavy computational overhead \cite{lakshminarayanan2017simple, durasov2021masksembles}. While effective in controlled settings, these approaches face trade-offs between scalability, accuracy, and efficiency, leaving room for alternative strategies \cite{mi2022training}.

A widely adopted strategy for uncertainty estimation leverages the relationship between reconstruction error and predictive reliability \cite{wang2025trustworthyvitalsignforecasting}. The intuition is straightforward: \emph{if an input can be faithfully reconstructed from its internal representation, it is likely to resemble the training distribution, and the model’s prediction is expected to be reliable. Conversely, large reconstruction errors often signal distributional shifts, which are associated with higher uncertainty}. Building on this principle, Reconstruction Uncertainty Estimation (RUE) \cite{korte2024confidence} has emerged as a representative approach. As illustrated in Figure~\ref{fig:intro2}, RUE defines uncertainty as the discrepancy between a query instance $\mathbf{x}$ and its reconstruction $\hat{\mathbf{x}}$. Instances closer to the training data should, in theory, exhibit smaller reconstruction errors and more reliable predictions, while inputs far from the training distribution tend to yield larger errors and higher uncertainty.

Despite its appeal, this strategy faces a fundamental limitation. Reconstruction error does not purely reflect uncertainty but also captures information lost during deep feature propagation. Even training samples may reconstruct imperfectly because intermediate layers inevitably discard fine-grained details unrelated to the predictive task. As a result, RUE can conflate genuine predictive uncertainty with artifacts of representation learning. 

This shortcoming is particularly evident in medical imaging. For example, in glaucoma detection, clinically relevant cues lie in the optic disc and cup, whereas reconstruction-based uncertainty maps often highlight vessels or peripheral regions (see Figure~\ref{fig:intro2}). Such misattribution arises because these structures are difficult to reconstruct, not because they contribute to diagnostic uncertainty. Addressing this entanglement between true uncertainty and information loss remains a key challenge for advancing reconstruction-based methods.

In this work, we propose Difference Reconstruction Uncertainty Estimation (DRUE). Instead of comparing a reconstruction directly with the input, DRUE introduces two reconstruction models attached to different intermediate layers of the predictive model and quantifies the discrepancy between their reconstructions as the uncertainty score. By leveraging differences between reconstructions at different depths, DRUE reduces the influence of cumulative information loss while preserving sensitivity to prediction-relevant signals.

The contributions of this work are as follows:

1. We propose DRUE, a novel approach for robust uncertainty estimation that mitigates the confounding effect of information loss in reconstruction-based methods.

2. We evaluate DRUE under varying degrees of distribution shift across multiple classification datasets, demonstrating consistent improvements in the accuracy of uncertainty estimation.

\section{Methodology}

\begin{figure}[!htb]
  \centering
  \includegraphics[width=\linewidth]{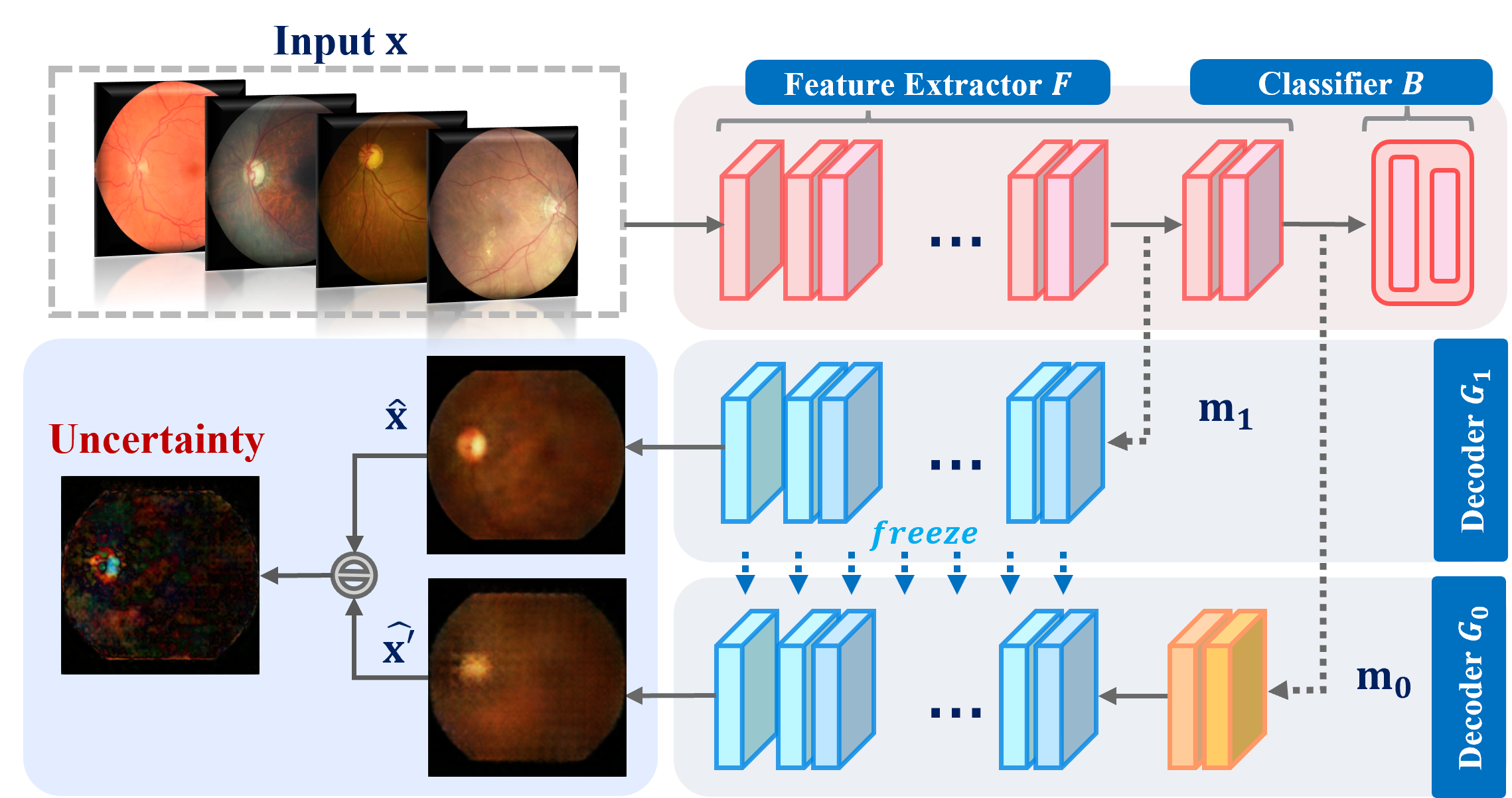}
  \caption{Pipeline of DRUE. The upper part represents the classification model $M$, which takes an input $\mathbf{x}$ and generates a prediction $\hat y$. Attached below are the two decoders $G_1$ and $G_0$.}
  \label{fig:method}
\end{figure}

\subsection{Connection between Reconstruction Error and Uncertainty}
We consider prediction tasks where a deep learning model $M$ maps an input $\mathbf{x} \in \mathbb{R}^i$ to an output $y$. During training, inputs are sampled from a distribution $p(\mathbf{x})$, but at test time the model may encounter inputs $\mathbf{x'}$ that differ due to noise or distribution shift. In such cases, estimating the reliability of predictions is essential, motivating the use of uncertainty estimation.  

Most deep learning models can be decomposed into a feature extractor $F$ and a predictor $B$. Beyond prediction, the features extracted by $F$ can be leveraged to reconstruct the input through an auxiliary decoder $G$. The reconstruction error, defined as the difference between the input $\mathbf{x}$ and its reconstruction $G(F(\mathbf{x}))$, has been studied as a proxy for uncertainty $U_R(\mathbf{x})$ \cite{wang2023uncertainty, korte2024confidence}. Formally:
\begin{align}
U_R(\mathbf{x}) = \|G(F(\mathbf{x})) - \mathbf{x}\|_1.
\end{align}
Larger reconstruction errors are often associated with less reliable predictions.  

However, reconstruction error does not exclusively reflect predictive uncertainty. A substantial part of the error can result from information loss during feature extraction, rather than from uncertainty itself. This relationship can be expressed as:
\begin{align}
U_R(\mathbf{x}) = U_{loss} + U_{real}, \label{EQ1}
\end{align}
where $U_{real}$ represents genuine predictive uncertainty and $U_{loss}$ arises from the loss of information between the input and deep features. For the output $\mathbf{m_l}$ of the $l$-th block, information loss can be written as:
\begin{align}
U_{loss} = H(\mathbf{x}) - I(\mathbf{x}, \mathbf{m_l}) = H(\mathbf{x}|\mathbf{m_l}).
\end{align}

This decomposition exposes a fundamental limitation: reconstruction error alone cannot disentangle information loss from true uncertainty. Reconstructing from deeper layers considers more task-relevant features but accumulates larger loss, while reconstructing from shallower layers preserves more information yet captures less predictive uncertainty. This trade-off reduces the precision and reliability of reconstruction-based uncertainty estimation, highlighting the need for improved approaches.

\begin{table*}[ht]
    \centering
    \caption{Performance of uncertainty estimation. The best-performing model for each metric has been highlighted in bold, and the next best has been underlined.}
    {\fontsize{9pt}{9pt}\selectfont
    \begin{tabular}{lcccccccc}
    \toprule 
    \multirow{2}{*}{\bfseries{Method}} & \multicolumn{2}{c}{\bfseries{\underline{PAPILA}}} & \multicolumn{2}{c}{\bfseries{\underline{ACRIMA}}} & \multicolumn{2}{c}{\bfseries{\underline{HAM10000}}} & \multicolumn{2}{c}{\bfseries{\underline{CIFAR10}}}\\
     &AUC($\uparrow$) &AUPR($\uparrow$) &AUC($\uparrow$) &AUPR($\uparrow$) &AUC($\uparrow$) &AUPR($\uparrow$) &AUC($\uparrow$) &AUPR($\uparrow$)\\
    \midrule 
    DRUE &\textbf{0.87 ± 0.04} &\textbf{0.79 ± 0.08} &\textbf{0.99 ± 0.00} &\textbf{0.99 ± 0.00} &\textbf{1.00 ± 0.00} &\textbf{1.00 ± 0.00} &\textbf{1.00 ± 0.00} &\textbf{1.00 ± 0.00}\\
    Entropy &0.38 ± 0.02 &0.33 ± 0.01 &0.49 ± 0.06 &0.46 ± 0.04 &0.80 ± 0.05 &0.84 ± 0.04 &0.88 ± 0.06 &\underline{0.98 ± 0.01}\\
    MC Dropout &0.64 ± 0.07 &0.54 ± 0.04 &\underline{0.85 ± 0.01} &\underline{0.80 ± 0.03} & \underline{0.90 ± 0.02} &\underline{0.93 ± 0.01} &\underline{0.94 ± 0.02} &\textbf{1.00 ± 0.00}\\
    PostNet Alea. & 0.65 ± 0.06 & 0.49 ± 0.06 & 0.57 ± 0.04 & 0.50 ± 0.04 & 0.57 ± 0.04 & 0.68 ± 0.03 & 0.90 ± 0.07 & 0.93 ± 0.06\\
    PostNet Epis. &0.58 ± 0.10 & 0.40 ± 0.09 & 0.70 ± 0.14 & 0.66 ± 0.14 & 0.68 ± 0.11 & 0.79 ± 0.08 & 0.83 ± 0.12 & 0.89 ± 0.08\\
    DEC &0.52 ± 0.02 &0.46 ± 0.02 &0.68 ± 0.00 &0.69 ± 0.01 &0.66 ± 0.03 &0.81 ± 0.02 &0.66 ± 0.02 &0.96 ± 0.00\\
    BNN &\underline{0.71 ± 0.04} &\underline{0.61 ± 0.05} &0.71 ± 0.07 &0.70 ± 0.04 &0.69 ± 0.10 &0.82 ± 0.05 &0.64 ± 0.11 &0.96 ± 0.01\\
    \bottomrule 
    \end{tabular}}
    \label{outtest}
\end{table*}

\subsection{Difference Reconstruction Uncertainty Estimate (DRUE)}
As shown in Figure~\ref{fig:method}, the DRUE is composed of two parts, Decoder $G_1$ and Decoder $G_0$. Decoder $G_0$ takes the output of the feature extractor $F$ as its input, while Decoder $G_1$ takes the feature extracted from the penultimate block of $F$ as input. We use uppercase letters for block combinations and lowercase letters for individual blocks. The feature extractor $F$, consisting of $l$ blocks, is represented as $F_l$, while the $l$-th block of the network is represented as $f_l$. 
\begin{definition}[DRUE uncertainty]
Let $\hat{\mathbf{x}}'$ and $\hat{\mathbf{x}}$ be the reconstructed outputs from decoders $G_0$ and $G_1$, respectively. The DRUE uncertainty $U_D$ is defined as the error between two reconstructed results: 
\begin{align} 
U_D(\mathbf{x}) = \|\hat {\mathbf{x}} - \hat{\mathbf{x}}'\|_1.
\end{align}
\end{definition}
The structure of Decoders mirrors the structure of the feature extractor. Decoders $G_0$ and $G_1$ share parameters after the first block of $G_0$. In other words, $G_1$ is a part of $G_0$. In the training stage, we first trained Decoder $G_1$ and then trained Decoder $G_0$ with frozen parameters loaded from $G_1$. Both decoders compute the MAE to estimate uncertainty. The loss function for the reconstruction task is formulated as:
\begin{align} 
L_{G_0} &= \frac{1}{n}\sum_{i=1}^n(x_i-G_0(F_l(x))_i)^2, \nonumber\\
L_{G_1} &= \frac{1}{n}\sum_{i=1}^n(x_i-G_1(F_{l-1}(x))_i)^2.
\end{align}

We establish the validity of DRUE from two complementary perspectives: {\em gradient-based reconstruction} and {\em information-theoretic analysis}, as follows.

\textbf{Gradient-based reconstruction.} First, we show that the difference between two reconstructed outputs, $\hat{\mathbf{x}}$ and $\hat{\mathbf{x}}'$, is equivalent to the sum of the reconstruction errors in the final block, weighted by gradient attributions. The features of the last and the penultimate block, $\mathbf{m_0}$ and $\mathbf{m_1}$, can be computed with the following equations:
\begin{align} 
  \mathbf{m_1} &= F_{l-1}(\mathbf{x}), \nonumber\\
  \mathbf{m_0} &= F_{l}(\mathbf{x}) = F_{l-1}(f_l(\mathbf{x})).
\end{align}
The reconstructed outputs $\hat{\mathbf{x}}$ and $\hat{\mathbf{x}}'$ from decoders $G_1$ and $G_0$ can also be represented as:
\begin{align} 
  \hat {\mathbf{x}} &= G_1(\mathbf{m_1}) = G_1(F_{l-1}(\mathbf{x})),\nonumber\\
  \hat {\mathbf{x}}' &= G_0(\mathbf{m_0}) = G_1(g_0(f_l(F_{l-1}(\mathbf{x})))),
\end{align}
where $g_0$ maps the final-layer features to the penultimate feature space. To compute the difference between $\hat{\mathbf{x}}$ and $\hat{\mathbf{x}}'$, we represent $F_{l-1}(\mathbf{x})$ as $\mathbf{z}$ and the difference $g_0(f_l(F_{l-1}(\mathbf{x}))) - F_{l-1}(\mathbf{x})$ as $\mathbf{\Delta z}$:
\begin{align} 
\hat {\mathbf{x}}' - \hat {\mathbf{x}} = G_1(\mathbf{z} + \mathbf{\Delta z}) - G_1(\mathbf{z}).
\end{align}
By applying a first-order Taylor expansion, we approximate the difference as follows:
\begin{align} 
\hat {\mathbf{x}}' - \hat {\mathbf{x}} &\approx J_{G_1}(\mathbf{z}) \cdot \Delta \mathbf{z} \nonumber \\
& = \frac{\partial G_1}{\partial z_0}\cdot \Delta z_0 + \frac{\partial G_1}{\partial z_1}\cdot \Delta z_1 + ... + \frac{\partial G_1}{\partial z_k}\cdot \Delta z_k,
\end{align}
where $J_{G_1}(\mathbf{z})$ is the Jacobian matrix of $G_1$ evaluated at $\mathbf{z}$, $z_0$ to $z_k$ are the elements of $\mathbf{z}$. Thus, the uncertainty $U_D(\mathbf{x})$ of the prediction is computed as:
\begin{align} 
U_D(\mathbf{x}) &\approx ||J_{G_1}(\mathbf{z}) \cdot \Delta \mathbf{z}||_1.
\end{align}

The DRUE uncertainty consists of two components: $\Delta z_0$, representing the reconstruction error of feature $m_1$, and the coefficient $\frac{\partial G_1}{\partial z_0}$, representing the derivative of $G_1$ with respect to $z_0$.  In other words, DRUE captures the effect of deep feature perturbations on the reconstruction output. Notably, the information loss in the final block is less than in the entirety of the feature extractor. By incorporating gradients, DRUE also reflects the uncertainty stemming from the feature extractor, ensuring both reduced information loss and accurate uncertainty estimation.

\textbf{Information-theoretic perspective.} Next, we validate DRUE from an information-theory standpoint. Suppose the parameters of $G_1$ are the same as the parameters after the first block of $G_0$. The mutual information between $\mathbf{x}$ and $\hat{\mathbf{x}}'$ quantifies how much information about $\mathbf{x}$ is preserved in the final output. The information loss is formulated as:
\begin{align} 
&H(\mathbf{x})-I(\mathbf{x},\hat {\mathbf{x}}')= H(\mathbf{x}|\hat {\mathbf{x}}')\nonumber\\
&= H(\mathbf{x} | G_1(g_0(f_l(F_{l-1}(\mathbf{x}))))).
\end{align}
Here, $H(\cdot)$ denotes Shannon entropy and $I(\cdot,\cdot)$ denotes mutual information. Similarly, the information loss from $\hat{\mathbf{x}}$ to $\hat{\mathbf{x}}'$ is computed as: 
\begin{align} 
&H(\hat{\mathbf{x}})-I(\hat {\mathbf{x}},\hat {\mathbf{x}}')= H(\hat {\mathbf{x}}|\hat {\mathbf{x}}') \nonumber\\
&= H(G_1(F_{l-1}(\mathbf{x})) | G_1(g_0(f_l(F_{l-1}(\mathbf{x}))))).
\end{align}
Note that for random variables $A$ and $B$ and a function $f$, the inequality $H(f(A) | B) \leq H(A | B)$ always holds, since any function process will introduce uncertainty \cite{klir1999uncertainty}. From this, we derive: 
\begin{align} 
 &H(\mathbf{x} | G_1(g_0(f_l(F_{l-1}(\mathbf{x}))))) \geq \nonumber\\
 &H(G_1(F_{l-1}(\mathbf{x})) | G_1(g_0(f_l(F_{l-1}(\mathbf{x}))))).
\end{align}
Thus, the information loss between the input $\mathbf{x}$ and the reconstructed result $\hat{\mathbf{x}}'$ is larger than the information loss between the input $\hat{\mathbf{x}}$ and the reconstructed result $\hat{\mathbf{x}}'$. This demonstrates DRUE’s ability to reduce information loss while maintaining its effectiveness in uncertainty estimation.

\section{Experiments}

\begin{figure*}[!htb]
  \centering
  \includegraphics[width=0.85\linewidth]{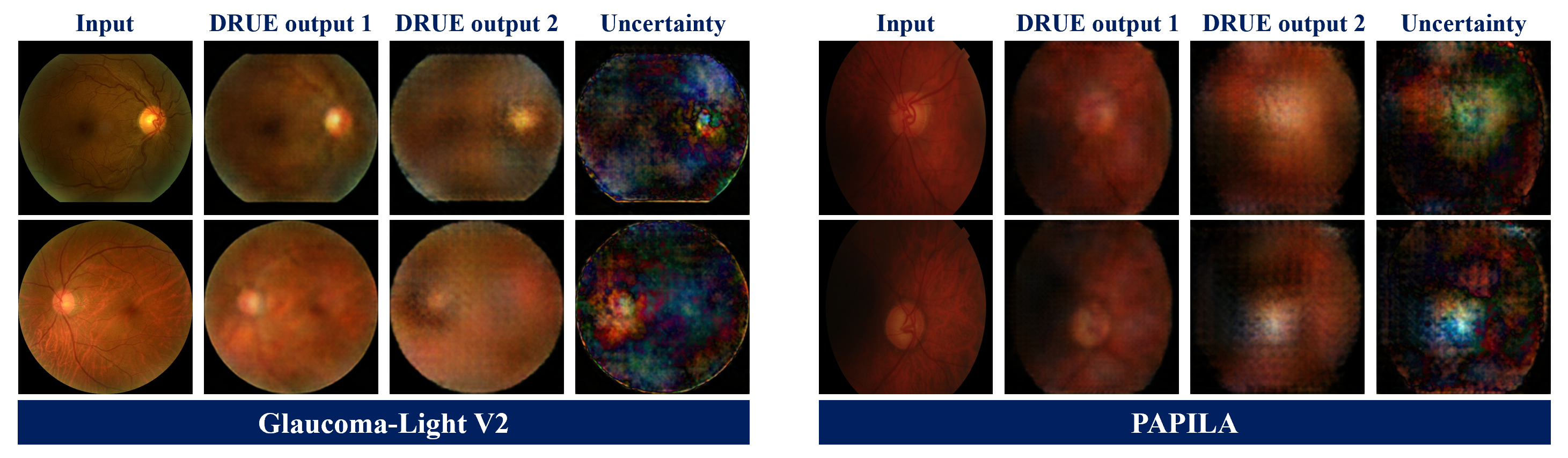}
  \caption{Visual results of input images, reconstructions from DRUE, and the corresponding uncertainty maps (normalized to [0, 1]).}
  \label{fig:vision}
\end{figure*}

\textbf{Experiment setting.}
OOD detection is a crucial application for uncertainty estimation \cite{wang2024epistemic}. We evaluate our method across five datasets that differ in domain, resolution, and image characteristics. Glaucoma-Light V2 \cite{kiefer2022survey} is used as the ID dataset due to its large size, balanced classes, and high image quality. Glaucoma-Light V2 is composed of 9,540 fundus images (4,770 RG / 4,770 NRG), split into 8,000 training, 770 validation, and 770 test samples. The OOD datasets—PAPILA \cite{kovalyk2022papila}, ACRIMA \cite{diaz2019cnns}, HAM10000 \cite{tschandl2018ham10000}, and CIFAR-10 \cite{krizhevsky2009learning}—represent progressively larger domain shifts from Glaucoma-Light V2. Specifically, PAPILA and ACRIMA are fundus datasets with different acquisition strategies and visual characteristics, HAM10000 comes from a distinct medical domain (dermoscopic skin images), and CIFAR-10 represents a non-medical natural image dataset. This progression allows us to assess uncertainty estimation under increasingly challenging shifts. 

In our experiments, all models were trained on Glaucoma-Light V2 and tested on all five datasets. The classification model and DRUE model are trained using the Adam optimizer with an initial learning rate of $1 \times 10^{-5}$, batch size of 4 and 8, and early stopping based on validation loss. We compared our DRUE model to existing methods: Entropy \cite{malinin2018predictive}, Monte Carlo Dropout (MC Dropout) \cite{folgoc2021mc}, PostNet \cite{charpentier2020posterior}, Deep Evidential Classification (DEC) \cite{sensoy2018evidential}, and BNN \cite{kendall2017uncertainties}. All models were trained with the best parameters found by random search and implemented using the ResNet18 architecture \cite{he2016deep} due to its proven efficiency and effectiveness in image classification tasks. The classification results are presented in Table \ref{perform}.

We quantify OOD detection performance using AUC and AUPR \cite{techapanurak2021practical}. AUC measures the ability to distinguish ID from OOD samples across thresholds, while AUPR emphasizes detection under class imbalance, providing a complementary evaluation.

\begin{table}
    \centering
    \caption{Mean accuracy and AUC of models serving as the basis for uncertainty estimation. For fairness, DRUE and entropy are evaluated on the same base classifier, while methods requiring re-training were optimized to achieve competitive classification performance.}
    {\fontsize{9pt}{10pt}\selectfont
    \begin{tabular}{lcc}
    \toprule 
    \bfseries Method & \bfseries AUC($\uparrow$) & \bfseries Accuracy($\uparrow$)\\
    \midrule 
    DRUE, Entropy & \textbf{0.97 ± 0.00}  &\underline{0.91 ± 0.00} \\
    MC Dropout & \textbf{0.97 ± 0.00} & \textbf{0.92 ± 0.01}\\
    PostNet Aleatoric & 0.85 ± 0.04  & 0.77 ± 0.03 \\
    PostNet Epistemic & 0.85 ± 0.04 & 0.77 ± 0.03 \\
    DEC & \underline{0.88 ± 0.00} & 0.88 ± 0.00	\\
    BNN & \textbf{0.97 ± 0.01} & \underline{0.91 ± 0.00} \\
    \bottomrule 
    \end{tabular}}
    \label{perform}
\end{table}

\begin{figure}[!htb]
  \centering
  \includegraphics[width=\linewidth]{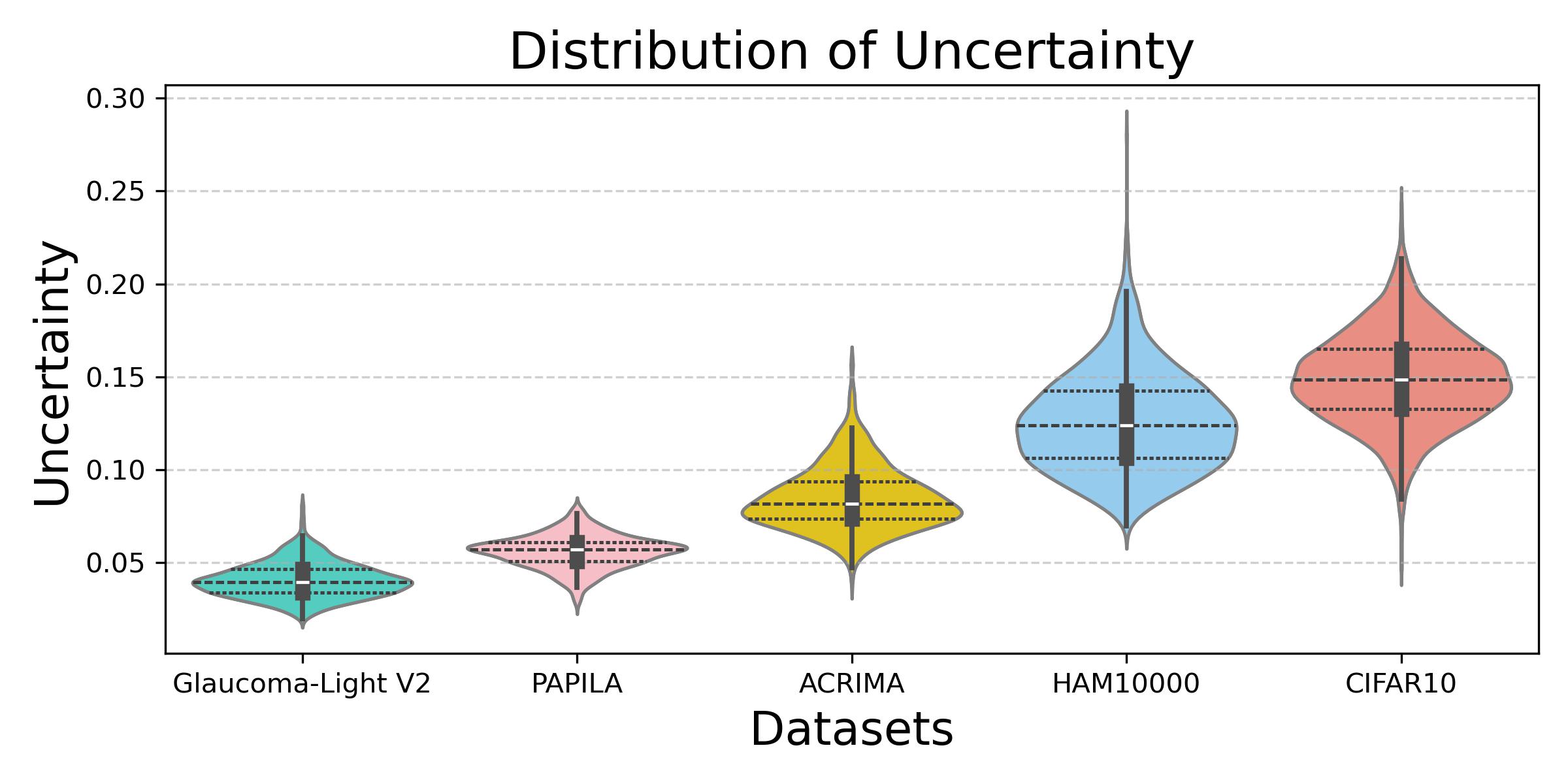}
  \caption{Distribution of DRUE uncertainty across datasets. Uncertainty increases with greater data shift.}
  \label{fig:disresult}
\end{figure}

\textbf{Quantitative evaluation of uncertainty estimation performance.} Table~\ref{outtest} summarizes the results across all datasets. DRUE consistently achieves the best performance. Among the baselines, BNN tends to follow DRUE on datasets with subtle domain shifts (PAPILA), suggesting that Bayesian neural networks are more effective when detecting small deviations from the training distribution. By contrast, MC Dropout performs comparatively better on larger shifts (CIFAR10), where strong distributional differences make stochastic forward passes more informative. Entropy-based measures remain unreliable, as softmax outputs are known to be overconfident~\cite{malinin2018predictive}, while DEC and PostNet struggle due to optimization difficulties and lower base model accuracy. Overall, these results demonstrate that DRUE provides a more accurate and stable framework for uncertainty estimation across diverse shift scenarios. Figure \ref{fig:disresult} shows the distribution of the uncertainty across all samples from the five datasets. The results align with the intuition that the uncertainty of the model’s predictions increases as the difference between the predicted data and the training data grows.

\textbf{Reconstruction visual results.} 
Figure \ref{fig:vision} illustrates the visual results of the reconstructions generated by DRUE on the glaucoma datasets (Glaucoma-Light V2 and PAPILA). Diagnosing glaucoma relies heavily on analyzing the size and shape of the Optic Disc (OD) and Optic Cup (OC), which are the bright, round regions in the center of retinal images \cite{armaly1969cup, kumar2019rim}. The figure illustrates that by comparing the outputs of two reconstruction models at different layers, DRUE provides accurate attribution by focusing on the areas truly relevant to the prediction, such as the optic disc and cup without vessels.

\textbf{Ablation study.} 
We conducted an ablation study to evaluate the contribution of the two-decoder structure on the PAPILA dataset, as shown in Table~\ref{abla}. The results indicate that using a single reconstruction model from either the feature extractor output (row 0) or the penultimate block (row 1, as $G_1$) yields lower performance, particularly in terms of AUPR. Incorporating the freeze strategy (row 2, as $G_0$) improves stability compared with $G_1$. The best results are achieved when combining both decoders (row 3), which consistently delivers the highest AUC (0.867) and similar AUPR (0.786). These findings confirm the complementary effect of the two-decoder design and the importance of mitigating information loss.

\begin{table}
    \centering
    \caption{Ablation study on PAPILA dataset.}
    {\fontsize{9pt}{10pt}\selectfont
    \begin{tabular}{lcccc}
    \toprule 
    &\bfseries Location &\bfseries Freeze  &\bfseries AUC($\uparrow$) &\bfseries AUPR($\uparrow$)\\
    \midrule 
    0 & 0 &  & \underline{0.86 ± 0.03} & \underline{0.76 ± 0.04}\\
    1 & 1 &  & 0.68 ± 0.03 & 0.50 ± 0.03\\
    2 & 0 & \Checkmark & 0.81 ± 0.08 & 0.68 ± 0.12\\
    3 & 0 + 1 & \Checkmark & \textbf{0.87 ± 0.04} & \textbf{0.79 ± 0.08}\\
    \bottomrule 
    \end{tabular}}
    \label{abla}
\end{table}

\section{Conclusion}
In this work, we present DRUE, a novel approach for improving uncertainty estimation in deep learning. By leveraging the reconstruction error between two models, DRUE reduces the influence of information loss and provides more reliable measures of uncertainty. Future directions include extending DRUE to regression tasks, exploring its applicability across diverse architectures, and investigating its potential for linking uncertainty to specific input features. Overall, this work advances uncertainty estimation by enhancing the accuracy and robustness of uncertainty assessment.

\vfill\pagebreak

\bibliographystyle{IEEEbib}
\bibliography{redsnew}

\end{document}